%% file: main.tex
\documentclass[11pt]{article}

\usepackage[margin=1in]{geometry}
\usepackage{amsmath,amssymb}
\usepackage{booktabs}
\usepackage{graphicx}
\usepackage{hyperref}
\usepackage{enumitem}
\usepackage{microtype}
\usepackage{natbib}
\usepackage{xcolor}

\hypersetup{
  colorlinks=true,
  linkcolor=blue,
  citecolor=blue,
  urlcolor=blue
}

\graphicspath{{figures/}}

\title{Continual Learning as Shared-Manifold Continuation Under Compatible Shift}
\author{Henry J. Kobs\\\texttt{henrykryon@gmail.com}}
\date{}

\newcommand{\method}{\textsc{SPMA-OG}}
\newcommand{\spma}{\textsc{SPMA}}

\begin{document}
\maketitle

\begin{abstract}
Continual learning methods usually preserve old behavior by regularizing parameters, matching old outputs, or replaying previous examples. These strategies can reduce forgetting, but they do not directly specify how the latent representation should evolve. We study a narrower geometric alternative for the regime where old and new data should remain on the same latent support: \emph{continual learning as continuation of a shared manifold}. We instantiate this view within Support-Preserving Manifold Assimilation (\spma{}) and evaluate a geometry-preserving variant, \method{}, that combines sparse replay, output distillation, relational geometry preservation, local smoothing, and chart-assignment regularization on old anchors. On representative compatible-shift CIFAR10 and Tiny-ImageNet runs, \method{} improves over sparse replay baselines in old-task retention and representation-preservation metrics while remaining competitive on new-task accuracy. On a controlled synthetic atlas-manifold benchmark, it achieves near-perfect anchor-geometry preservation while also improving new-task accuracy over replay. These results provide evidence that geometry-aware anchor regularization is a useful inductive bias when continual learning should preserve a shared latent support rather than create a new one.
\end{abstract}

\section{Introduction}
Catastrophic forgetting remains a central difficulty in continual learning \citep{delange2022survey}. When a model is fine-tuned on new data, the latent representation often changes in ways that are globally disruptive for previously learned behavior. Existing methods usually operate at one of three levels: parameter preservation \citep{kirkpatrick2017ewc,zenke2017si,aljundi2018mas}, functional preservation through distillation \citep{hinton2015distilling,li2017lwf}, or replay of previous examples \citep{lopez2017gem,chaudhry2019agem,rolnick2019experience,buzzega2020dark}. These strategies can be effective, but they do not directly encode a geometric preference over how the representation itself should move.

We study a representation-level hypothesis: in many continual-learning settings, the new task should not require a completely new latent topology. Instead, new data should be absorbed into the \emph{same shared manifold} already used by the old task. The goal is therefore neither to freeze coordinates exactly nor to allocate fully orthogonal task subspaces. Rather, we aim to preserve the global support and geometry of the old manifold while permitting only local, task-necessary deformation.

We study this hypothesis through the \spma{} family. The paper's empirical focus is \method{}, a geometry-preserving instantiation of \spma{} that models the old representation as a collection of local latent charts fit to frozen teacher features. During fine-tuning, old anchors preserve chart structure and relational geometry, while new-task adaptation is driven by task supervision under those same anchor-based geometric constraints. The resulting bias favors shared-support continuation over global reorganization.

The contribution is therefore not a claim of broad continual-learning superiority, but a targeted result about compatible-shift settings where old and new samples plausibly share latent support. Our falsifiable expectation is that, in this regime, \method{} should improve representation preservation and old-task retention with only modest new-task cost relative to sparse replay. We view that scoped geometric framing, together with the resulting empirical tradeoff, as the primary contribution.

\section{Related Work}
Continual-learning methods are commonly organized into parameter-regularization, replay, and distillation families \citep{delange2022survey}. Parameter-based approaches such as EWC, Synaptic Intelligence, and MAS estimate parameter importance and penalize updates to important weights \citep{kirkpatrick2017ewc,zenke2017si,aljundi2018mas}. These methods directly address forgetting at the parameter level, but they do not explicitly constrain how the representation geometry should evolve.

Replay remains one of the strongest practical paradigms. Gradient-constrained replay methods such as GEM and A-GEM preserve past performance through episodic memories and constrained updates \citep{lopez2017gem,chaudhry2019agem}. Simpler rehearsal variants and retrieval-based methods also remain competitive, including experience replay, MIR, and DER++ \citep{rolnick2019experience,aljundi2019mir,buzzega2020dark}. In class-incremental recognition, replay is often coupled with distillation, as in iCaRL and LUCIR \citep{rebuffi2017icarl,hou2019lucir}.

Our work is most closely related to representation-preserving continual learning. Relational Knowledge Distillation preserves inter-example geometry rather than matching coordinates directly \citep{park2019rkd}. PODNet distills pooled intermediate structure for class-incremental learning \citep{douillard2020podnet}. Backward Feature Projection and related methods constrain how features may change under fine-tuning \citep{zheng2023bfp}. Subspace Distillation emphasizes stable low-dimensional structure during continual adaptation \citep{li2023subspace}. We build on these ideas but adopt a more explicit geometric interpretation: the old representation is treated as a collection of local latent charts, and continual learning is posed as continuation of that shared manifold rather than preservation of coordinates alone.

\section{Method}
\subsection{Problem setup}
Let $f_{\theta_0}$ be a model trained on an old task, and let $f_\theta$ be a student initialized from $f_{\theta_0}$ and fine-tuned on a new task. Let $h(\cdot)$ denote a chosen latent representation layer. We assume access to:
\begin{itemize}[leftmargin=1.25em]
  \item new-task data $D_{\text{new}}$,
  \item a small old-task anchor set $D_{\text{anc}}$,
  \item and frozen teacher features $z_0(x)=h_{\theta_0}(x)$.
\end{itemize}
The goal is to learn the new task while preserving the old latent manifold globally and allowing only local, task-necessary deformation.

\subsection{Local chart memory}
We build a compact memory from teacher features on $D_{\text{anc}}$ by clustering $\{z_0(x)\}$ into $K$ local components. For each component $k$, we estimate a low-rank factor model
\[
z \approx \mu_k + U_k a + \varepsilon,
\qquad
a \sim \mathcal{N}(0,\Lambda_k),
\qquad
\varepsilon \sim \mathcal{N}(0,\sigma_k^2 I),
\]
where $\mu_k$ is the local mean, $U_k$ is an orthonormal basis for the local tangent subspace, $\Lambda_k$ contains retained factor variances, and $\sigma_k^2$ models isotropic residual variation. The corresponding covariance is $\Sigma_k = U_k \Lambda_k U_k^\top + \sigma_k^2 I$. In the experiments, chart memory is built once from frozen teacher features using $K$-means and local low-rank fits; it is not refreshed during fine-tuning.

This chart memory serves two roles. First, it provides a coarse atlas of the old manifold. Second, it defines a soft chart assignment for each feature vector through factor-model scores. For a feature $z$, the score of chart $k$ is
\[
s_k(z) =
\frac{1}{d}
\left(
\sum_{r=1}^{r_k}\frac{a_{k,r}(z)^2}{\lambda_{k,r}+\sigma_k^2}
\;+\;
\frac{\lVert z - \mu_k - U_k a_k(z)\rVert_2^2}{\sigma_k^2}
\;+\;
\log \det \Sigma_k
\right),
\]
where $d$ is the feature dimension, $r_k$ is the retained rank, and $a_k(z)=U_k^\top(z-\mu_k)$. We convert these scores into soft chart assignments with
\[
p_k(z)=\frac{\exp(-s_k(z)/\tau_c)}{\sum_j \exp(-s_j(z)/\tau_c)}.
\]

\subsection{SPMA objective family}
The broader \spma{} family combines task supervision with output preservation, geometric regularization, chart preservation, and optional continuation constraints on new-task samples:
\begin{equation}
\begin{aligned}
\mathcal{L}_{\mathrm{SPMA}}
=\;&
\mathcal{L}_{\text{new}}
+ \lambda_{\text{KD}} \mathcal{L}_{\text{KD}}
+ \lambda_{\text{anchor}} \mathcal{L}_{\text{anchor}}
+ \lambda_{\text{geo}} \mathcal{L}_{\text{geo}} \\
&+ \lambda_{\text{smooth}} \mathcal{L}_{\text{smooth}}
+ \lambda_{\text{chart}} \mathcal{L}_{\text{chart}}
+ \lambda_{\text{cont}} \mathcal{L}_{\text{cont}} \\
&+ \lambda_{\text{support}} \mathcal{L}_{\text{support}}
+ \lambda_{\text{reg}} \mathcal{L}_{\text{reg}}.
\end{aligned}
\end{equation}
The continuation and support terms act directly on new-task samples by penalizing departures from the old chart atlas. They are part of the broader \spma{} viewpoint, but the experiments in this paper evaluate a simpler geometry-preserving instance, \method{}, which omits those explicit new-sample terms. We therefore treat the full objective as conceptual context and \method{} as the concrete method under study.

\subsection{The \method{} instantiation}
Let $B_{\text{new}}$ denote a new-task batch and $B_{\text{anc}}$ an old-anchor replay batch. The \method{} objective is
\begin{equation}
\begin{aligned}
\mathcal{L}_{\mathrm{SPMA\text{-}OG}}(t)
=\;&
\mathcal{L}_{\text{new}}
+ \beta(t)\lambda_{\text{anchor}}\mathcal{L}_{\text{anchor}} \\
&+ \alpha(t)\Big(
\lambda_{\text{KD}}\mathcal{L}_{\text{KD}}
+ \lambda_{\text{geo}}\mathcal{L}_{\text{geo}}
+ \lambda_{\text{smooth}}\mathcal{L}_{\text{smooth}} \\
&\qquad\qquad
+ \lambda_{\text{chart}}\mathcal{L}_{\text{chart}}
+ \lambda_{\text{reg}}\mathcal{L}_{\text{reg}}
\Big).
\end{aligned}
\end{equation}
where $\alpha(t)$ and $\beta(t)$ are linearly decayed retention schedules over fine-tuning.

The individual losses are:
\begin{align*}
\mathcal{L}_{\text{new}}
&=
\frac{1}{|B_{\text{new}}|}
\sum_{(x,y)\in B_{\text{new}}}
\mathrm{CE}(f_\theta(x), y), \\
\mathcal{L}_{\text{anchor}}
&=
\frac{1}{|B_{\text{anc}}|}
\sum_{(x,y)\in B_{\text{anc}}}
\mathrm{CE}(f_\theta(x), y), \\
\mathcal{L}_{\text{KD}}
&=
\frac{T^2}{|B_{\text{anc}}|}
\sum_{x\in B_{\text{anc}}}
\mathrm{KL}\!\left(
\sigma\!\left(\frac{f_{\theta_0}(x)}{T}\right)
\middle\|
\sigma\!\left(\frac{f_{\theta}(x)}{T}\right)
\right).
\end{align*}

To preserve geometry, we compute normalized pairwise distance matrices on anchor features,
\[
\widetilde{D}_{ij}(z) = \frac{\lVert z_i - z_j\rVert_2}{\text{mean off-diagonal Euclidean distance}},
\]
and match them with
\[
\mathcal{L}_{\text{geo}}
=
\frac{1}{m(m-1)}
\sum_{i\neq j}
\left(
\widetilde{D}_{ij}(z) - \widetilde{D}_{ij}(z_0)
\right)^2,
\]
where $m=|B_{\text{anc}}|$.

We additionally preserve local geometry through a teacher $k$-nearest-neighbor graph:
\[
\mathcal{L}_{\text{smooth}}
=
\frac{\sum_{i,j} w_{ij}
\left(
\widetilde{D}_{ij}(z) - \widetilde{D}_{ij}(z_0)
\right)^2}
{\sum_{i,j} w_{ij}},
\qquad
w_{ij}=\exp\!\left(-\widetilde{D}_{ij}(z_0)/\tau_s\right),
\]
where the weights are restricted to teacher-neighbor pairs. This term penalizes local tearing of the old manifold.

Finally, chart preservation matches the teacher and student soft chart assignments on old anchors:
\[
\mathcal{L}_{\text{chart}}
=
\tau_c^2
\frac{1}{|B_{\text{anc}}|}
\sum_{x\in B_{\text{anc}}}
\mathrm{KL}\!\big(
p(z_0(x)) \,\|\, p(z(x))
\big),
\]
and $\mathcal{L}_{\text{reg}}$ is an $\ell_2$ penalty on parameter drift relative to $\theta_0$.

Operationally, $\mathcal{L}_{\text{anchor}}$ preserves supervised old-task behavior on sparse replay, $\mathcal{L}_{\text{KD}}$ preserves teacher logits, $\mathcal{L}_{\text{geo}}$ preserves global anchor geometry, $\mathcal{L}_{\text{smooth}}$ preserves local neighborhoods, $\mathcal{L}_{\text{chart}}$ discourages wholesale reassignment of anchors to different charts, and $\mathcal{L}_{\text{reg}}$ limits excessive parameter drift. The new-task cross-entropy then adapts the model under these anchor-based constraints, encouraging continuation of the old manifold rather than global representational drift.

\begin{figure}[t]
  \centering
  \includegraphics[width=\linewidth]{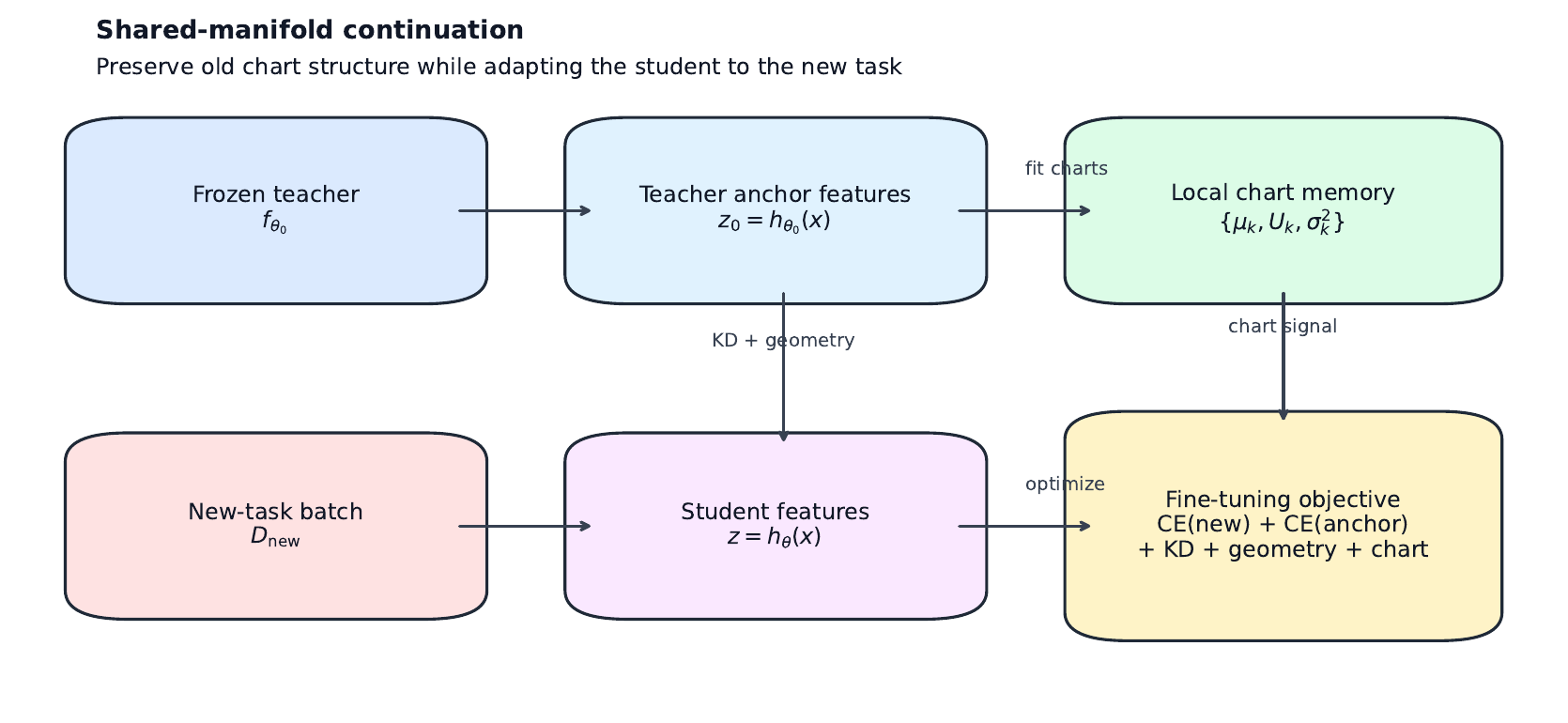}
  \caption{Overview of the \spma{} viewpoint. Frozen teacher features define a local chart memory. Old anchors preserve function, geometry, and chart structure; the student is then adapted to the new task using cross-entropy together with anchor-based geometric regularization.}
  \label{fig:method}
\end{figure}

\section{Experimental Setup}
\subsection{Benchmarks}
We focus on two compatible-shift settings that are directly aligned with the shared-manifold hypothesis:
\begin{enumerate}[leftmargin=1.5em]
  \item \textbf{CIFAR10 compatible shift}: the same CIFAR10 classes under a shifted input distribution.
  \item \textbf{Tiny-ImageNet compatible shift}: same semantic classes under a new shifted input distribution.
\end{enumerate}
Both settings keep the same semantic classes and were chosen deliberately as hypothesis tests, not as claims about general class-incremental learning. In both benchmarks, the new input distribution applies the same family of image transformations used in the saved runs: a $12^\circ$ rotation, translation by $(+2,-1)$ pixels, scale 0.95, additive noise with standard deviation 0.03, and a blur kernel of size 3.

\subsection{Models and protocol}
For CIFAR10 we use a ResNet-18 adapted to CIFAR; for Tiny-ImageNet we use an ImageNet-style ResNet-18 backbone. Old-task teachers are trained first and then frozen. Fine-tuning uses a single shared classifier because the label space does not change between old and new data. The representative runs reported here use seed 7, 10 base-training epochs, and 5 fine-tuning epochs on both benchmarks. Anchor memories are built once from frozen teacher features, with 64 buffered anchors per class and cluster-stratified replay for sparse-anchor methods. The full old anchor set contains 640 examples on CIFAR10 and 12{,}800 on Tiny-ImageNet; ER-512 and \method{} both replay 512 anchors, while Anchor CE replays 256. \method{} uses the same sparse replay budget as ER-512, but adds linearly decayed retention terms for KD, geometry, smoothing, and chart preservation.

\subsection{Baselines}
We compare against:
\begin{itemize}[leftmargin=1.25em]
  \item \textbf{Plain FT}: standard fine-tuning.
  \item \textbf{Anchor CE}: sparse supervised replay on old anchors.
  \item \textbf{ER-512}: sparse replay with 512 old anchors.
  \item and \textbf{SPMA-OG}: the old-geometry preserving SPMA variant that serves as the main method in this paper.
\end{itemize}

\subsection{Metrics}
We report old-task accuracy after fine-tuning, new-task accuracy after fine-tuning, forgetting, harmonic mean of old and new accuracy, linear CKA, pairwise-distance correlation, and empirical support inclusion. Because some variants use different support score parameterizations, we use CKA and pairwise-distance correlation as the primary cross-method representation metrics and treat support inclusion as a secondary diagnostic.

\section{Results}
\subsection{Compatible-shift benchmarks}
Table~\ref{tab:main_results} summarizes the CIFAR10 and Tiny-ImageNet compatible-shift runs used throughout the paper.

\begin{table}[t]
  \centering
  \caption{Fixed-seed compatible-shift results on CIFAR10 and Tiny-ImageNet.}
  \label{tab:main_results}
  \footnotesize
  \resizebox{\linewidth}{!}{\input{tables/main_results.tex}}
\end{table}

These benchmarks are deliberately chosen to test the regime in which old and new data should remain on the same latent support. On CIFAR10 compatible shift, \method{} is best in these representative runs: it retains old-task accuracy at $0.8195$ versus $0.7948$ for ER-512 while also reaching higher new-task accuracy ($0.7906$ vs.\ $0.7755$). On Tiny-ImageNet compatible shift, \method{} retains substantially more old-task accuracy than ER-512 ($0.3059$ vs.\ $0.2218$) while matching new-task accuracy ($0.3250$ vs.\ $0.3254$). Taken together, these runs are consistent with the intended hypothesis: when the new data should lie on the same broad support, stronger geometry preservation need not require large new-task sacrifice.

\subsection{Representation preservation}
The stronger case for \method{} is representational rather than purely leaderboard-based. Table~\ref{tab:repr_results} compares selected methods on CKA, pairwise-distance correlation, and empirical support inclusion.

\begin{table}[t]
  \centering
  \caption{Representation-level diagnostics for the same representative runs. \method{} improves substantially over replay-only in preserving old structure while remaining competitive on new-task accuracy.}
  \label{tab:repr_results}
  \footnotesize
  \resizebox{\linewidth}{!}{\input{tables/representation_results.tex}}
\end{table}

On CIFAR10 compatible shift, \method{} improves over ER-512 from 0.8966 to 0.9209 in CKA and from 0.8670 to 0.8970 in pairwise-distance correlation. On Tiny-ImageNet compatible shift, the gain is much larger: CKA rises from 0.4439 to 0.7525 and pairwise-distance correlation from 0.3962 to 0.7773. These diagnostics are consistent with the paper's main claim: the full geometry-preserving package yields a more stable old representation than replay alone, not just a different old/new accuracy tradeoff.

\subsection{Synthetic manifold sanity check}
To isolate the geometry claim more directly, we also evaluate on a controlled synthetic atlas-manifold benchmark in which a warped ribbon surface is observed through two different input maps while the semantic classes are kept fixed. We treat this as a corroborating sanity check rather than a main empirical pillar.

\begin{figure}[t]
  \centering
  \includegraphics[width=\linewidth]{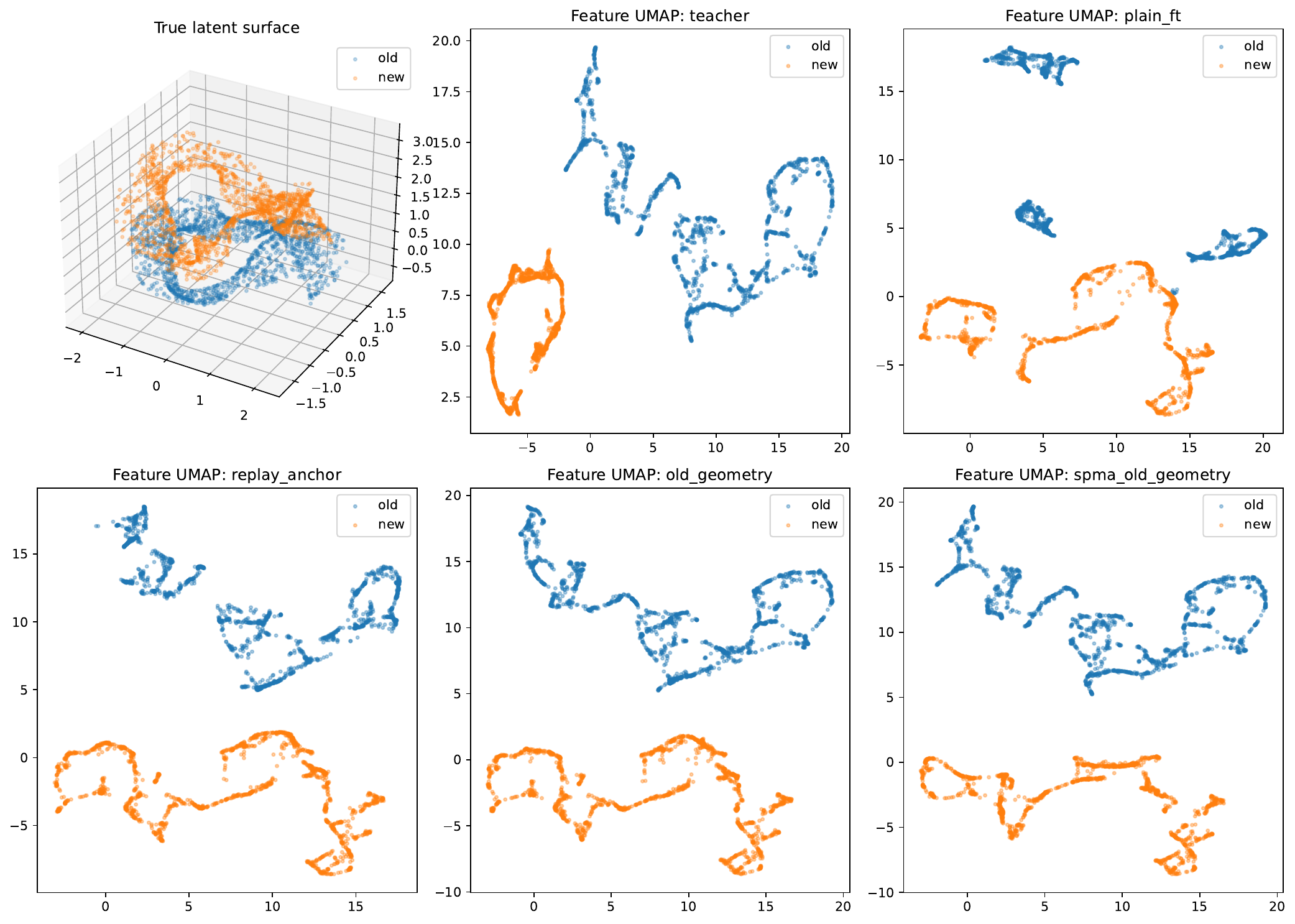}
  \caption{Synthetic atlas-manifold benchmark. The true latent surface is shown at left, followed by teacher, plain fine-tuning, replay-anchor, old-geometry, and \method{} feature views. \method{} preserves the old manifold most faithfully while still absorbing the new task into the same shared support.}
  \label{fig:synthetic_overview}
\end{figure}

\begin{table}[t]
  \centering
  \caption{Synthetic atlas-manifold results. \method{} achieves the best old-task accuracy, the best new-task accuracy among the manifold-preserving methods, and near-perfect anchor geometry preservation.}
  \label{tab:synthetic_results}
  \footnotesize
  \resizebox{\linewidth}{!}{\input{tables/synthetic_results.tex}}
\end{table}

On this synthetic manifold, \method{} improves over replay-anchor from $0.9213$ to $0.9269$ on old-task accuracy and from $0.8712$ to $0.8875$ on new-task accuracy, while raising anchor CKA from $0.9813$ to $0.9994$ and anchor pairwise-distance correlation from $0.9759$ to $0.9992$. This controlled result is consistent with the geometric interpretation and shows near-perfect preservation on the anchor-based geometry metrics used in the paper.

\begin{figure}[t]
  \centering
  \includegraphics[width=\linewidth]{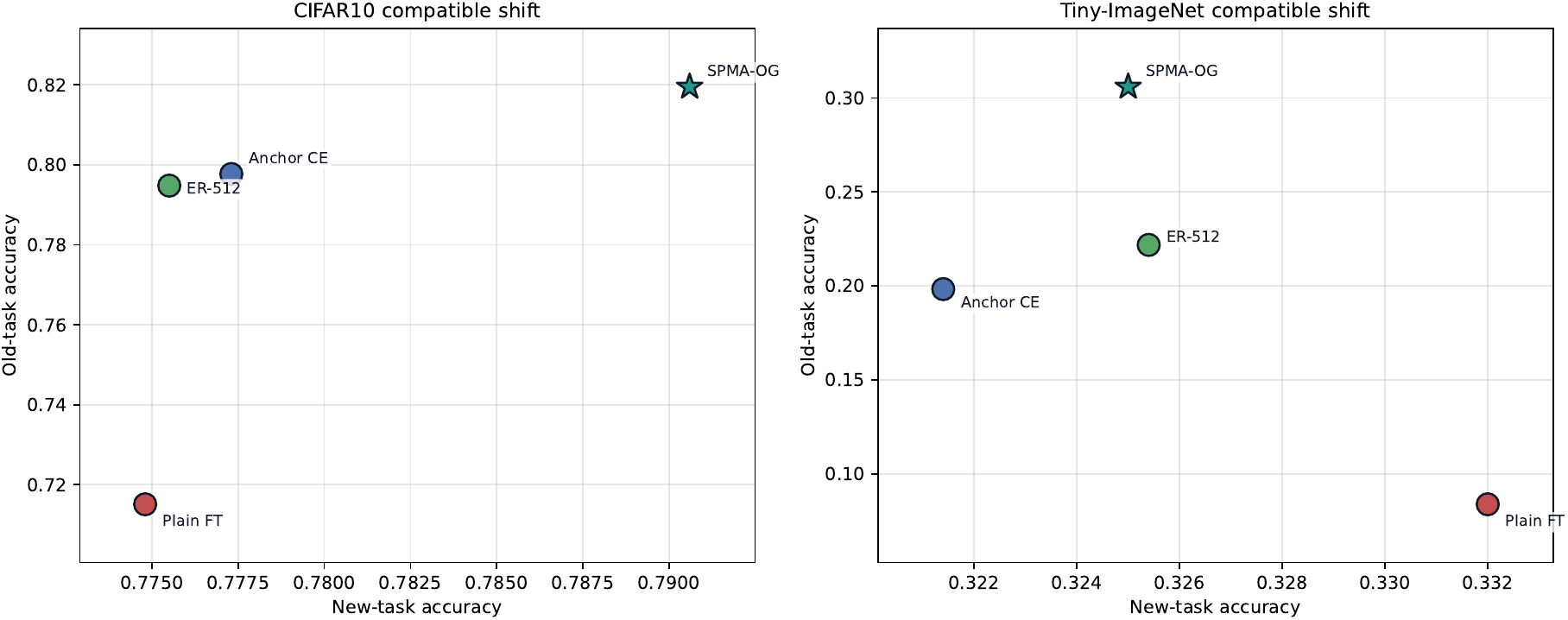}
  \caption{Old/new tradeoff on the two compatible-shift benchmarks used in the paper. \method{} occupies the best or near-best region while preserving the strongest latent geometry.}
  \label{fig:tradeoff}
\end{figure}

\begin{figure}[t]
  \centering
  \includegraphics[width=\linewidth]{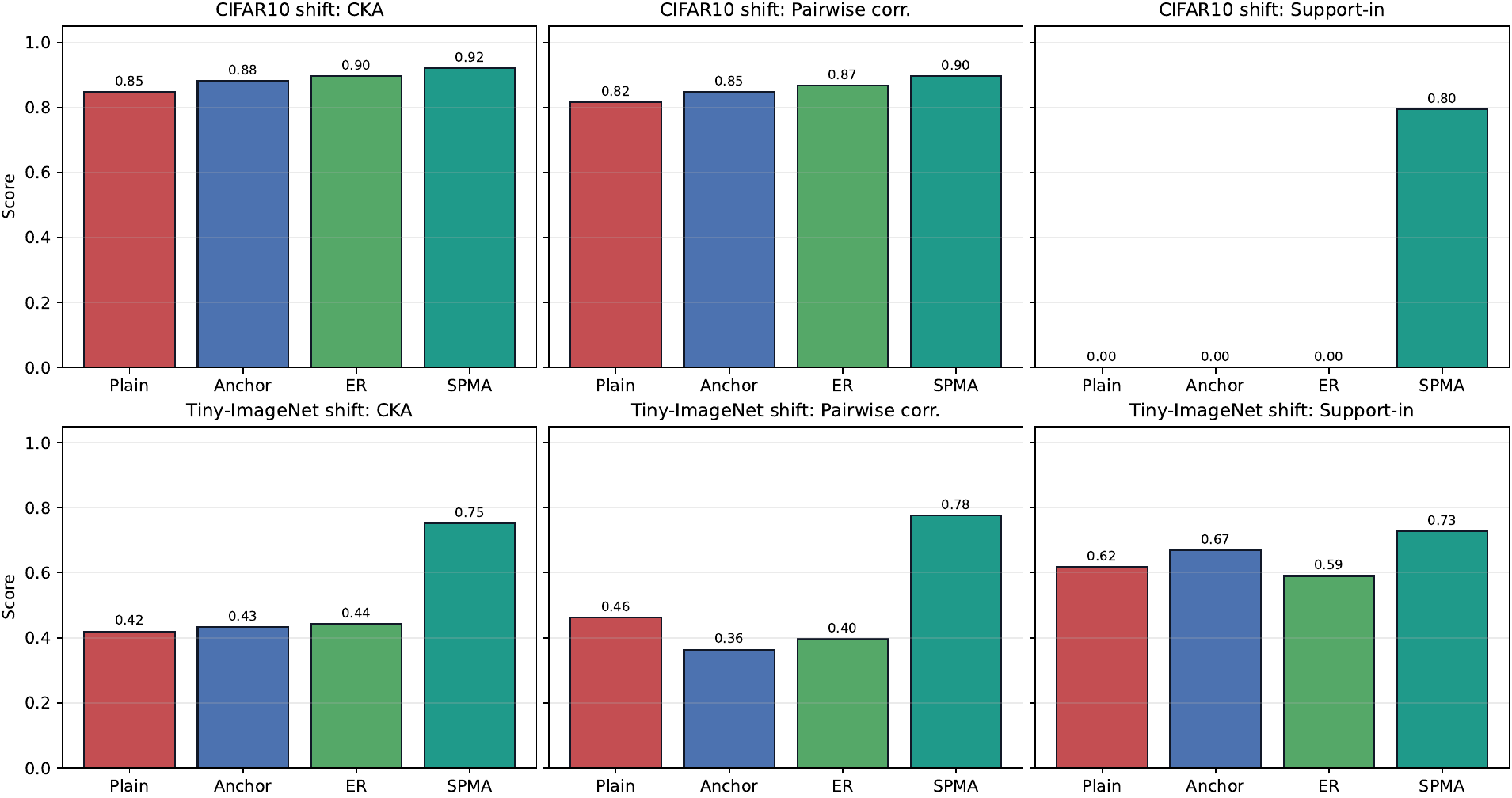}
  \caption{Representation preservation summary. Compared with replay-only and plain fine-tuning, \method{} yields markedly stronger CKA and pairwise-distance correlation, especially on Tiny-ImageNet compatible shift.}
  \label{fig:repr}
\end{figure}

\subsection{Detailed Tiny-ImageNet run}
The Tiny-ImageNet compatible-shift benchmark is the setting most closely aligned with the original manifold-assimilation hypothesis. Table~\ref{tab:shift_results} reports that run directly.

\begin{table}[t]
  \centering
  \caption{Tiny-ImageNet compatible-shift run. We report old-task accuracy both before and after fine-tuning to make the same-class setting explicit.}
  \label{tab:shift_results}
  \footnotesize
  \resizebox{\linewidth}{!}{\input{tables/compatible_shift_results.tex}}
\end{table}

The teacher starts at $0.3482$ old-task accuracy, and \method{} retains $0.3059$ after fine-tuning while reaching $0.3250$ on the shifted new data. We view this as the most direct large-scale evidence in the paper for the intended shared-support behavior, because it keeps the semantic label space fixed while perturbing only the input distribution.

\section{Discussion}
The main contribution of \method{} is not that it universally dominates all baselines on task accuracy. Rather, it offers a coherent third operating point between unrestricted drift and rigid freezing in the compatible-shift regime studied here.

On the compatible-shift benchmarks, replay remains a strong task-level baseline. However, replay alone often preserves old behavior without preserving shared latent structure, whereas \method{} retains substantially more old-manifold structure and, on both shift benchmarks, better old-task accuracy as well.

We therefore view this work as a representation-centric continual-learning result. The current evidence supports the full geometry-preserving package of replay, distillation, geometry, smoothing, chart preservation, and drift control; it does not isolate chart preservation alone as the sole source of the gains.

The paper also does not establish that shared-manifold continuation is the right inductive bias for generic class-incremental or novel-class continual learning. When new tasks require genuine representational expansion, the shared-support assumption may be too restrictive. The claim is therefore intentionally narrower: geometry-aware anchor regularization appears useful when old and new data should plausibly inhabit the same latent support.

\section{Conclusion}
We introduced the \spma{} viewpoint and evaluated the geometry-preserving variant \method{} in compatible-shift settings where old and new data plausibly share latent support. Instead of freezing old features or allocating orthogonal task subspaces, the method preserves old chart structure while adapting to new data under explicit anchor-based geometric constraints. Across the CIFAR10 and Tiny-ImageNet compatible-shift benchmarks, \method{} remains competitive with sparse replay on task accuracy while showing substantially stronger representation preservation. On the synthetic atlas-manifold benchmark, it also preserves the old geometry more faithfully than replay alone. These results do not establish a general continual-learning state of the art, but they do provide evidence that shared-manifold continuation is a useful framing in the regime it is intended to model.

\bibliographystyle{plainnat}
\bibliography{references}

\end{document}

%% file: tables/main_results.tex
\begin{tabular}{lrrrrrrrr}
\toprule
Method & Replay & CIFAR10-Shift Old & CIFAR10-Shift New & CIFAR10-Shift Harm. & Tiny-Shift Old & Tiny-Shift New & Tiny-Shift Harm.\\
\midrule
Plain FT & 640 & 0.7151 & 0.7748 & 0.7438 & 0.0838 & \textbf{0.3320} & 0.1338\\
Anchor CE & 256 & 0.7977 & 0.7773 & 0.7874 & 0.1983 & 0.3214 & 0.2453\\
ER-512 & 512 & 0.7948 & 0.7755 & 0.7850 & 0.2218 & 0.3254 & 0.2638\\
SPMA-OG & 512 & \textbf{0.8195} & \textbf{0.7906} & \textbf{0.8048} & \textbf{0.3059} & 0.3250 & \textbf{0.3152}\\
\bottomrule
\end{tabular}

%% file: tables/representation_results.tex
\begin{tabular}{lrrrrrr}
\toprule
Method & CIFAR10-Shift CKA & CIFAR10-Shift Dist. Corr. & CIFAR10-Shift Support-In & Tiny CKA & Tiny Dist. Corr. & Tiny Support-In\\
\midrule
Plain FT & 0.8485 & 0.8166 & 0.0000 & 0.4200 & 0.4625 & 0.6183\\
Anchor CE & 0.8816 & 0.8476 & 0.0000 & 0.4330 & 0.3639 & 0.6702\\
ER-512 & 0.8966 & 0.8670 & 0.0000 & 0.4439 & 0.3962 & 0.5907\\
SPMA-OG & \textbf{0.9209} & \textbf{0.8970} & \textbf{0.7950} & \textbf{0.7525} & \textbf{0.7773} & \textbf{0.7279}\\
\bottomrule
\end{tabular}

%% file: tables/synthetic_results.tex
\begin{tabular}{lrrrrr}
\toprule
Method & Old After & New After & Harmonic Mean & Anchor CKA & Anchor Dist. Corr.\\
\midrule
Plain FT & 0.5800 & \textbf{0.8994} & 0.7052 & 0.9563 & 0.9470\\
Replay Anchor & 0.9213 & 0.8712 & 0.8956 & 0.9813 & 0.9759\\
Old Geometry & 0.9169 & 0.8719 & 0.8938 & 0.9883 & 0.9834\\
SPMA-OG & \textbf{0.9269} & 0.8875 & \textbf{0.9068} & \textbf{0.9994} & \textbf{0.9992}\\
\bottomrule
\end{tabular}

%% file: tables/compatible_shift_results.tex
\begin{tabular}{lrrrrr}
\toprule
Method & Replay & Old Before & Old After & New After & Harmonic Mean\\
\midrule
Plain FT & 12800 & \textbf{0.3482} & 0.0838 & \textbf{0.3320} & 0.1338\\
Anchor CE & 256 & \textbf{0.3482} & 0.1983 & 0.3214 & 0.2453\\
ER-512 & 512 & \textbf{0.3482} & 0.2218 & 0.3254 & 0.2638\\
SPMA-OG & 512 & \textbf{0.3482} & \textbf{0.3059} & 0.3250 & \textbf{0.3152}\\
\bottomrule
\end{tabular}